\RequirePackage{fix-cm}

\documentclass[11pt,a4paper]{article}

\usepackage[utf8]{inputenc}
\usepackage{graphicx}
\usepackage{pgfplots}
\usepackage{natbib}
\usepackage{subfig}
\usepackage{csquotes}
\usepackage{hyperref}
\pgfplotsset{height=8cm,width=12cm}

\begin{document}

\title{FinnSentiment – A Finnish Social Media Corpus for Sentiment Polarity Annotation}

\author{Krister Lindén \and
        Tommi Jauhiainen \and
        Sam Hardwick     
}

\maketitle

University of Helsinki\\
\{krister.linden,tommi.jauhiainen,sam.hardwick\}@helsinki.fi

\begin{abstract}
Sentiment analysis and opinion mining is an important task with obvious application areas in social media, e.g. when indicating hate speech and fake news. 
In our survey of previous work, we note that there is no large-scale social media data set with sentiment polarity annotations for Finnish. 
This publications aims to remedy this shortcoming by introducing a 27,000 sentence data set annotated independently with sentiment polarity by three native annotators. 
We had the same three annotators for the whole data set, which provides a unique opportunity for further studies of annotator behaviour over time.
We analyse their inter-annotator agreement and provide two baselines to validate the usefulness of the data set.

\end{abstract}

\section{Introduction}
Sentiment analysis and opinion mining is an important task that has its roots in market analysis where the market sentiment may determine the direction of the stock market. 
Customer sentiment analysis has gained traction in commercial product and brand name monitoring. 
With an increasing use of social media, opinion mining has an obvious application area in indicating hate speech and fake news.

An abundance of research has gone into sentiment analysis and data set creation in various languages. 
In Section~\ref{sec:previouswork}, we give a brief overview of the situation in various languages. 
Lately, shared tasks in sentiment research have become common place in various data genres.
In Section~\ref{sec:prevfinsent}, we in particular look at research using or creating Finnish sentiment data and find that a large data set for Finnish social media polarity sentiment data is lacking.

To remedy the situation, we picked 100,000 random sentences from a leading Finnish social media site -- \emph{Suomi24}.
A brief inspection found that most of the data would likely be neutral, so to save manual annotation time we bootstrapped the procedure by creating two external methods for indicating the likely sentiment of all the sentences as described in Section~\ref{sec:preannotation}. 
We picked half of the data to be manually annotated from sentences with sentiment indications that both methods agreed on and the rest of the data from the remaining portions of the data set. 
We selected 27,000 sentences and divided them into 9 work packages as outlined in Section~\ref{sec:corpus}.

We used three Finnish native speakers as manual annotators of the data set. 
After an initial training session, the annotators were instructed to work individually.
They all received the same data packages with 3,000 sentences for which to indicated a positive, negative or neutral sentiment.
All work packages were completed by all three annotators, and in Section~\ref{sec:annotationanalysis}, we analyse the individual sentiment indications over time as well as their mutual agreement. 
We also take a closer look at some examples on which the annotators disagree. 
As described in Section~\ref{sec:dataset}, based on the annotator indications for each sentence in the data set, we provide the majority vote and a derived 5-grade sentiment scale often used in shared tasks. 
We split the data into 20-folds for performing cross-validation. Finally, we describe the file format in which each sentence and the scores are provided.

To demonstrate the usefulness of the data set, we perform two baseline experiments with the data set in Section~\ref{sec:experiments}. 
We use one lexicon-based method, which is independent of the data set, and one neural network based model, which we train on our data set and use cross-validation for testing. 
We also perform some initial analysis of where the models diverge from the human analysis, and conclude the paper with a discussion and conclusion in Section~\ref{sec:discussionconclusion}.

\section{Previous work}
\label{sec:previouswork}

For an introduction to sentiment analysis, we refer the reader to the survey by \citeauthor{pang2008} in 2008. 
Their work was followed by \cite{liu2012}, who gives an in-depth introduction to sentiment analysis and opinion mining and presents a comprehensive survey of all important research topics up until 2012.

\cite{feldman2013} reviews some of the main research questions for sentiment analysis.
In 2014, \cite{medhat2014} surveyed the algorithms and applications for sentiment analysis. 
Their intention was to update the earlier work and give newcomers a panoramic view of the field. 
They also categorize the available benchmark data sets at the time. Later, also \cite{ravi2015} give a survey on opinion mining and sentiment analysis and list publicly available data sets known to them.

Later additions to surveys concerning sentiment analysis have been made by \cite{giachanou2016}, who discuss sentiment analysis for Twitter, and by \cite{zhang2018}, who make a survey of deep learning techniques used in sentiment analysis. 
\cite{mantyla2018evolution} present a computer-assisted review of the evolution of sentiment analysis analyzing 6,996 papers from Scopus.

Out of this vast number of articles, we have ourselves collected information concerning some of the previously published sentiment annotated data sets. 
These lists function as an easy access point to data sets available for different languages.\footnote{The newest data set descriptions for a language usually refer to the older ones and are findable, e.g. using Google Scholar.}
In Section~\ref{ssec:langspecdatasets}, we describe language specific data sets and in Section~\ref{ssec:multilangdatasets} multilingual data sets.

\subsection{Language specific data sets for sentiment analysis}
\label{ssec:langspecdatasets}

In this Section, we first introduce data sets for English as English has the largest variation. We then briefly mention data sets in other languages in alphabetical order according to language.

\cite{bostan2018analysis} compare several \textbf{English} emotion corpora in a systematic manner. They perform cross-corpus classification experiments using the available data sets, some of which are mentioned next.
\cite{wiebe2005} describe a 10,000-sentence corpus in English annotated with opinions, emotions, sentiments, etc. From these sentences, they annotated ``direct subjective frames", 1689 in total, with values \emph{positive} (8\%), \emph{negative} (23\%), \emph{both} ($<$1\%), or \emph{neither} (69\%). The corpus was collected as part of the Multi-Perspective Question Answering (MPQA) workshop. \cite{deng2015} present an entity/event-level annotation scheme for the MPQA 2.0 corpus \citep{wilson2008} and use it to create the MPQA 3.0 corpus.  
For the SemEval-2007 Task 14: Affective Text, \cite{strapparava2007semeval} made available a manually annotated set of 1,250 news headlines.
\cite{kessler2010} describe a large data set consisting of blog posts in English annotated with sentiment expressions, among others.  
\cite{maas2011} introduce a data set of 50,000 movie reviews from IMDB. 
\cite{saif2013evaluation} present a manually annotated data set of tweets called STS-Gold. 
The tweets for the data set were randomly collected from a larger corpus of tweets.
As part of the Concept Level Sentiment Analysis challenge, \cite{recupero2014} introduce a manually labeled data set in English. 
The data set includes 2,322 sentences tagged either as negative or positive. 
\cite{takala2014} annotated a collection of financial news samples from the Thomson Reuters news wire. The collection included 297 documents and over 9,000 sentences.
\cite{liu-etal-2019-dens} introduce DENS, a data set for emotions of narrative sequences. 
They used Amazon Mechanical Turk to crowd-source the annotations of Plutchik's eight core emotions \citep{plutchik1980general}.
\cite{demszky2020goemotions} published a manually annotated data set of 58,000 Reddit comments using 27 emotion categories in addition to neutral.

\cite{abdulla2014} present a manually annotated data set for \textbf{Arabic}. 

\cite{ku2007} annotated a set of documents from the NTCIR CIRB020 and CIRB040 traditional \textbf{Chinese} test collections \citep{sasaki2007}. Each sentence of the 843 documents was tagged with information relating to 32 different topics and in case of opinions, their polarities. All together 11,907 sentences were tagged. Each sentence was annotated by three annotators with respect to being \emph{positive}, \emph{negative}, or \emph{neutral}. 
\cite{ku2010} constructed the Chinese Opinion Treebank. It contains 18,782 sentences, which are annotated as positive, neutral or negative in addition to other annotations.

\cite{apidianaki2016} describe data sets for aspect-based sentiment analysis in \textbf{French}. 

\cite{clematide2012} describe a publicly available reference corpus for sentiment analysis in \textbf{German}.
 
\cite{boland2013} introduce an annotated data set for German product reviews. 
\cite{ruppenhofer2014} present the German sentiment analysis data set used in the GESTALT shared task. 

\cite{szabo2016} present a manually annotated sentiment corpus for \textbf{Hungarian}. 

\cite{bosco2013} and \cite{bosco2015} present Senti-TUT, a corpus of tweets in \textbf{Italian} annotated with sentiment polarity. 
They used five separate tags in annotation: positive, negative, ironic, mixed, and objective (neutral). \cite{basile2014} present the data set used in Evalita 2014 SENTIment POLarity Classification Task and \cite{barbieri2016} and \cite{basile2016} present the 2016 edition of the task. 
The task focused on sentiment classification of Italian tweets. 
\cite{stranisci2016} present a linguistic resource for sentiment analysis in Italian. 

\cite{tokuhisa2008emotion} collected 1.3 million \textbf{Japanese} texts from the Internet using an emotion lexicon and lexical patterns.

\cite{shin2012} and \cite{shin2013} describe the work of constructing a sentiment corpus for \textbf{Korean}.
\cite{jang2013} introduces a sentiment analysis corpus for Korean. 
\cite{chae2016} describe the methodology for constructing MUSE, a sentiment-annotated corpora for Korean created from the social-web. 

\cite{velldal2018norec} describe the \textbf{Norwegian} Review Corpus (NoReC) for training and evaluating models for document-level sentiment analysis. \cite{maehlum2019annotating} present a Norwegian data set for fine-grained sentiment analysis.

\cite{hosseini2018} describe a sentiment analysis corpus for \textbf{Persian}. 
The corpus containing more than 26,0000 sentences is annotated on the document-, sentence-, and entity/aspect-level.

\cite{carvalho2011} introduce \emph{SentiCorpus-PT}, a corpus of 2,795 online news comments comprising approx. 8,000 sentences in \textbf{Portuguese}.  
\cite{arruda2015} describe a corpus in Brazilian Portuguese annotated with paragraph polarity.

\cite{koltsova2016} describe a publicly available \textbf{Russian} test collection with sentiment markup and a crowd-sourcing website for such markup.
\cite{rogers2018} present RuSentiment, a data set for sentiment analysis of social media posts in Russian.

\cite{navas2019} provide a survey of \textbf{Spanish} corpora for sentiment analysis.

\cite{grubenmann2018} describe the \textbf{Swiss German} SB-CH corpus with sentiment annotations.

\cite{cuo2017} introduce TSTD, a data set for \textbf{Tibetan} sentiment analysis consisting of 10,000 tweets classified as positive, negative, and neutral.

\cite{omurca2017} describe a \textbf{Turkish} sentiment analysis corpus annotated at sentence level. 

\subsection{Multilingual data sets for sentiment analysis}
\label{ssec:multilangdatasets}

In this Section, we first present some multilingual sentiment analysis related shared tasks and their data sets. Then, we list some other multilingual data sets in chronological order grouped by the domain of the texts.

\cite{seki2007,seki2008,seki2010} give overviews of the \textbf{Opinion Analysis Pilot Tasks at the NT-CIR Workshops}. 

\cite{nakov2013,nakov2016} and \cite{rosenthal2014,rosenthal2015,rosenthal2017} describe the data sets used in \textbf{SemEval Sentiment Analysis in Twitter} tasks 2013-2017. 
\cite{nakov2016b} describe how the data sets were created for the 2013-2015 editions of the task.
\cite{pontiki2014,pontiki2015,pontiki2016} describe training and test data sets for the \textbf{SemEval task: Aspect Based Sentiment Analysis} 2014-2016. 
In 2016, there were 39 data sets for 8 languages.
\cite{ghosh2015} introduce the data set used in \textbf{SemEval-2015 Task 11: Sentiment Analysis of Figurative Language in Twitter}. 

\cite{hanig2014} present PACE, a multilingual evaluation corpus for phrase-level sentiment analysis. 
The corpus contains 2,000 posts from English and German \textbf{Internet forums}.

\cite{klinger2014} introduce the Bielefeld University Sentiment Analysis Corpus for German and English (USAGE) containing \textbf{product reviews} from Amazon. 
\cite{jimenez2015} present a manually annotated multi-lingual data set of hotel reviews. 

\cite{uryupina2014} present SenTube, a data set of \textbf{YouTube comments} annotated with sentiment polarity. 
The corpus includes English, Italian, Spanish, and Dutch. 

\cite{roman2015} describe an annotated corpus for \textbf{dialogue summaries} in English and Portuguese.

\cite{rei2016} introduce a multilingual \textbf{Twitter corpus} which includes annotation on sentiment polarity on the message-level. 
These tweets in German, Italian, and Spanish were also annotated with Part-of-Speech and Named Entity information.  

\section{Previous research in or using Finnish language sentiment analysis}
\label{sec:prevfinsent}

In this Section, we present the previous research and data sets used on sentiment analysis for Finnish.

Tiia Leuhu's master's thesis \citep{leuhu2014} is the earliest work we have identified that discusses automatic sentiment analysis for the Finnish language. 
Tweets in Finnish were manually annotated so that the collection consisted of 700 tweets in each of the three categories: positive, neutral, or negative. 
Using 10\% of the data for testing, she evaluated three machine learning algorithms: k-nearest neighbor, multinomial naïve Bayes, and random forest. 
Naïve Bayes proved to be the best algorithm for sentiment classification attaining the accuracy of 0.84. 
The annotated data set was not published.

\cite{paavola2015} used sentiment analysis in order to detect trolling behavior in tweets in Finnish during the 2014 Ukrainian crisis. 
They used a social media analysis tool developed in the NEMO project detecting the polarity (positive-neutral-negative) of the messages. The tool uses pre-defined positive and negative words and emoticons together with decision-tree and logic regression. 
The work was continued by \cite{paavola2016understanding,paavola2016automated} who analyzed Finnish tweets during the Syrian refugee crisis in order to detect bots. 
The tool is not currently available.

\cite{ohman2016} used the NRC Word-Emotion Association Lexicon \citep{mohammad2013crowdsourcing} to study the preservation of sentiments in translation in the Opensubtitles parallel corpus of movie subtitles \citep{lison2016opensubtitles2016} as well as the Europarl corpus in OPUS \citep{tiedemann2012parallel}. 
The word-emotion association lexicon was used to label sentences with one of the eight core emotions from Plutchik's wheel \citep{plutchik1980general} in addition to being generally negative or positive. 
The language pairs investigated were English - Finnish, English - Swedish, and Spanish - Portuguese. 
Using manually annotated sentences, they found that the Spanish - Portuguese pair has a higher cross-language agreement than the other two pairs.

\cite{ohman2018b} and \cite{ohman2018} introduce a web-based annotation tool called \emph{Sentimentator}.\footnote{https://github.com/Helsinki-NLP/sentimentator} 
Sentimentator uses a ten-dimensional model based on Plutchik's core emotions. 
Annotating sentences using a ten dimensional scheme requires more reflection from the annotator than simply tagging the sentence as positive, negative, or neutral. 
The authors set out to solve this by gamifying the process. In order to avoid domain bias, they set out to annotate the texts at sentence-level without a larger context as suggested by \cite{boland2013}. 
They used the Opensubtitles data set from OPUS with an initial focus on English and Finnish.

\cite{kajava2018} and \cite{kajava2020emotion} investigated sentiment preservation in translations and transfer learning.
Continuing the utilization of the Opensubtitles corpus, they used English sentences as the source and their Finnish, French, and Italian translations as targets. 
Each sentence was labeled with one of the Plutchik's core emotions using the Sentimentator annotation tool \citep{ohman2018b}. 
Once labeled, the English sentences were exported from Sentimentator and manually revised by a native English speaker who removed ambiguous or neutral sentences from the data set. 
The translations of the remaining sentences in Finnish, French, and Italian were similarly annotated by competent speakers, two for each language, and labeled with exactly one of the core emotions according to the speakers own judgment. 
The categorization of each sentence as negative or positive was then derived from these labels. 
In total, the data set consists of 6,427 sentences for each language. 
Cohen's Kappa coefficient was used as a measure for inter-annotator reliability \citep{cohen1960coefficient}. 
The sentiment preservation accuracy between English sentences and the translated sentences ranged from 0.82 for Italian to 0.86 for Finnish, indicating that sentiment is quite well preserved in translations. 
\cite{kajava2018} also created an evaluation data set with training and testing partitions and evaluated four machine learning classification algorithms: multilayer perceptron (MLP), multinomial naïve Bayes (MNB), support vector machine (LinearSVC), and maximum entropy (MaxEnt).\footnote{The data is available at https://github.com/cynarr/MA-thesis/tree/master/data-raw} 
Depending on the language, the best classification results were given by MNB, LinearSVC, or MaxEnt classifiers. 

\cite{ohman2020challenges} presents a continuation of the work using Sentimentator and the OPUS Movie Subtitle parallel corpus to annotate individual subtitle lines with Plutchik's core sentiments. 
She especially focuses on describing and evaluating the annotation process in detail. 
The result of the annotation work was over 56,000 annotated sentences in Finnish, Swedish, or English by roughly 100 separate annotators. 
\cite{ohman2020xed} published the XED data set with 25,000 Finnish and 30,000 English sentences annotated with Plutchik's core emotions.\footnote{https://github.com/Helsinki-NLP/XED} 
This is the largest data set release from the Helsinki-based research group so far continuing the work with Sentimentator \citep{ohman2018b} and open movie subtitle data from OPUS \citep{tiedemann2012parallel}. 
In addition to Finnish and English, the release includes projected annotations for 30 other languages. 
\cite{ohman2020c} is currently preparing manually verified versions of Finnish sentiment and emotion lexicons originally published by \cite{mohammad2013crowdsourcing}.

\cite{jussila2017} investigated the reliability of two sentiment analysis tools for Finnish when compared with human evaluators. 
The two analysis tools were the SentiStrength \citep{thelwall2010sentiment,thelwall2012sentiment} and the Nemo Sentiment and Data Analyzer \citep{paavola2015}. The Nemo Sentiment and Data Analyzer tool can also be used to collect tweets and it was used to collect a set of 509 tweets in Finnish. 
Two human annotators independently classified each of the tweets as positive, negative, or neutral. 
The Nemo Sentiment Analyzer can use one out of two separate algorithms to analyze sentiments: logistic regression and random forest. 
The SentiStrength returns the strength of positive and negative sentiment of the text on a scale from one to five. 
The values given by the three algorithms were used to classify the tweets as positive, negative, or neutral. 
The automatic classifications were then compared with the classifications of two human annotators. 
They used Krippendorff's alpha \citep{krippendorff2011computing} for evaluating the inter-annotator agreement and reliability of the annotations. The annotated data set was not published.

\cite{kaustinen2018} used a Finnish data set with 14,332 movie reviews rated from 1 to 10. The data was gathered from leffatykki.com on November 2017. He investigated what kind of effect linguistic differences between English and Finnish have on sentiment analysis.

\sloppy
\cite{nukarinen2018} used deep learning, Long Short-Term Memory (LSTM) recurrent neural networks, in experimenting with sentiment analysis in Finnish. 
For his experiments, he gathered over 50,000 product reviews from www.verkkokauppa.com. 
When classifying into categories from one to five, his classifier achieved an overall accuracy of 53.6\%.

\cite{einolander2019} analyzed textual customer feedback from Telia Finland. 
Several classification models were compared and a deep learning model utilizing LSTM networks performed the best.

\cite{vankka2019} implemented polarity lexicons for Finnish. 
They used reviews written in Finnish from the Trustpilot and TripAdvisor websites. 
The reviews were rated with values from 1 to 5. 
They created a hybrid algorithm using the polarity lexicons together with word embeddings. 
They found that using the headlines of the reviews instead of their content was less noisy as the content often describes both negative and positive sides of the reviewed item. 
The corpus they used is not currently available.

According to our review, currently the only available Finnish language data sets with manual sentiment annotations are those published by \cite{kajava2018} and \cite{ohman2020xed} based on movie subtitles.

\section{Preliminary Sentiment Annotations}
\label{sec:preannotation}

Prior to our current work, we implemented a CNN sentence classifier \citep{kim2014} for classifying texts for sentiment polarity, and trained this architecture on two data sets: a collection of product reviews scraped from online web stores, and sentences from the \emph{Suomi24} corpus containing emoticons. 
Emoticons were used as distant supervision similar to \cite{read2005using}, \cite{PAK10.385}, and \cite{abdul2017emonet}. 
We pretrained word embeddings for the model with \verb+word2vec+ \citep{mikolov1}.

\subsection{Product Review-based Annotator}

The product reviews contained a review text and a star rating, from $1$ to $5$ stars, reflecting total product satisfaction. 
We mapped this rating to a three-way sentiment classification by assigning $3$ as neutral, $< 3$ as negative and $> 3$ as positive.

\subsection{Smiley-based Annotator}

We took the intentionally naïve approach of directly taking a very limited interpretation of smileys as cues of sentiment in sentences. 
Those texts containing only positive smileys were assessed as positive, texts containing only negative smileys were assessed as negative and texts containing neither were assessed as neutral. 
Texts containing both positive and negative smileys were entirely discarded.

\subsection{Applying the pre-annotators}

These tools were initially tested by external users, but their reliability were deemed rather low. 
For some tasks like psychological priming experiments, the analyzer based on product reviews was felt to correlate better with human evaluations. 
This lead us to embark on a more extensive manual effort to annotate social media sentences with sentiment polarity. 
However, despite some social media discussions being inflamed, much of the text is still rather neutral, so to use the human annotation effort efficiently, we decided that the preliminary sentiment analyzers could be used to weed out some of the neutral sentences and raise the odds that there was at least a considerable number of sentences with sentiment polarity in the data to be given to the human annotators.

\section{Corpus}
\label{sec:corpus}

The original corpus consists of sentences from the social media site \emph{Suomi24}\footnote{www.suomi24.fi} which is available as a corpus through the Language Bank of Finland. 
From this corpus we randomly selected sentences and pre-annotated them with the pre-annotators for screening purposes. Based on the pre-annotations, we composed a corpus that was likely to have a higher proportion of non-neutral sentences which were annotated by human annotators for sentiment polarity.

\subsection{Text Selection Procedure}

First, we built a pre-selection corpus of $100,000$ random sentences from the \emph{Suomi24} corpus (data set release 2017H2 by \cite{suomi24-2001-2017-korp-v1-1_en}), without filtering on the basis of length or other criteria. 

We pre-evaluated our sample with our two automatic annotators, \emph{Product review} and \emph{Smiley}, and selected the sentences for human evaluation based on this pre-evaluation. 
The sentences in the pre-selected corpus were classified by the automated annotators as shown in Table~\ref{tab:preselected_distribution}.

\begin{table}[h]
\centering
\begin{tabular}{|r|c|c|c|}
\hline
& \multicolumn{3}{c|}{Smiley} \\
& \textbf{POS} & \textbf{NEUTR} & \textbf{NEG} \\
\hline
\textbf{POS} & 4,861 & 24,984 & 895 \\
Product review \textbf{NEUTR} & 3,007 & 18,914 & 1,891  \\
\textbf{NEG} & 4,494 & 35,274 & 5,680 \\
     \hline
\end{tabular}
\caption{Distribution of pre-selected sentences}
\label{tab:preselected_distribution}
\end{table}

The automated pre-evaluation annotators completely agreed on 29,455 sentences, slightly disagreed (one was neutral and the other was not) on 65,156 sentences and strongly disagreed on 5,389 sentences. 

This pre-selection corpus was then divided into four categories, which were used for selection into the final corpus in desired proportions. Well aware that annotating may sometimes be a time-consuming task, we also wanted to divide the work into work packages for the human annotators to let them feel that they had made visible progress when a work package had been completed. 
In each work package of $3,000$ sentences, we included sentences evaluated by both our automated pre-evaluation annotators, of which
\begin{itemize}
  \item $500$ had an agreed on positive sentiment,
  \item $500$ had an agreed on neutral sentiment,
  \item $500$ had an agreed on negative sentiment, and
  \item $1,500$ on which the automated annotators disagreed
\end{itemize}

As a result, the sentiment corpus of $27,000$ sentences had $4,500$ sentences with potentially positive, neutral, and negative evaluations each, respectively, and $13,500$ sentences with potentially no clear sentiment polarity. 
The corpus with potentially enriched polarity data had the distribution shown in Table~\ref{tab:selected-distrib}.

\begin{table}
\centering
\begin{tabular}{|r|c|c|c|}
\hline
& \multicolumn{3}{c|}{Smiley} \\
& \textbf{POS} & \textbf{NEUTR} & \textbf{NEG} \\
\hline
\textbf{POS} & 4,500  & 4,797  & 170 \\
Product review \textbf{NEUTR} & 573 & 4,500 & 356 \\
\textbf{NEG} & 869 & 6,735 & 4,500 \\
     \hline
\end{tabular}
\caption{Distribution of selected sentences}
\label{tab:selected-distrib}
\end{table}

The $27,000$ sentences comprised a total of $346,937$ tokens and $2,052,900$ (Unicode) characters, which is an average of $12.8$ tokens per sentence and $76$ characters per sentence.

\subsection{Annotators and annotation schema}

The annotators were students of language technology at the University of Helsinki. 
They were, however, unaccustomed to sentiment annotation, and we determined that in the interest of being able to obtain a sufficiently large corpus in a reasonable amount of time, it would be best to perform only a three-way annotation: positive, negative and neutral. 
Following \cite{boland2013} and \cite{ohman2018b} we decided that the sentences would be annotated without context.

\subsection{Annotation Process}

We assigned the 9 work packages of $3,000$ sentences to each of our annotators. 
As described, each package contained the same distribution of sentences from our pre-selection categories, but the sentences within each package were randomly shuffled, i.e. the sentences from each category did not appear consecutively. 
The work packages given to each annotator were identical.

After a brief initial meeting, the annotators worked independently of each other. 
They used a spreadsheet program to input their single-character annotation in column A for the sentence in column B.

There was no schedule set except a final deadline, and the bulk of the annotations was performed closer to the deadline than the beginning of the project.

To kick off the annotation task, we invited the annotators to a briefing. 
We described the task and advised the annotators that human agreement in this task is normally in the 70\% range. 
We explained that since the sentences were being presented out of context, it would not always be possible to judge the intended sentiment accurately, but they should avoid \emph{overthinking} and make a quick judgement call as to whether sentiment was either explicitly present or overwhelmingly likely in context.

After some discussion, the annotators did a trial run of 100 sentences to make sure they had some shared understanding of the task. 
We went over these annotations together.

\section{Analysis of the Annotations}
\label{sec:annotationanalysis}

To see how well the annotation schema was adhered to and how the perception of sentiment may vary between individuals, we look at the overall distribution of sentiment ratings, make an overview of the annotations by individual for each sentence in the corpus over time, and finally look at some annotated examples where the annotators totally agreed or disagreed.

\subsection{Distribution of annotations}

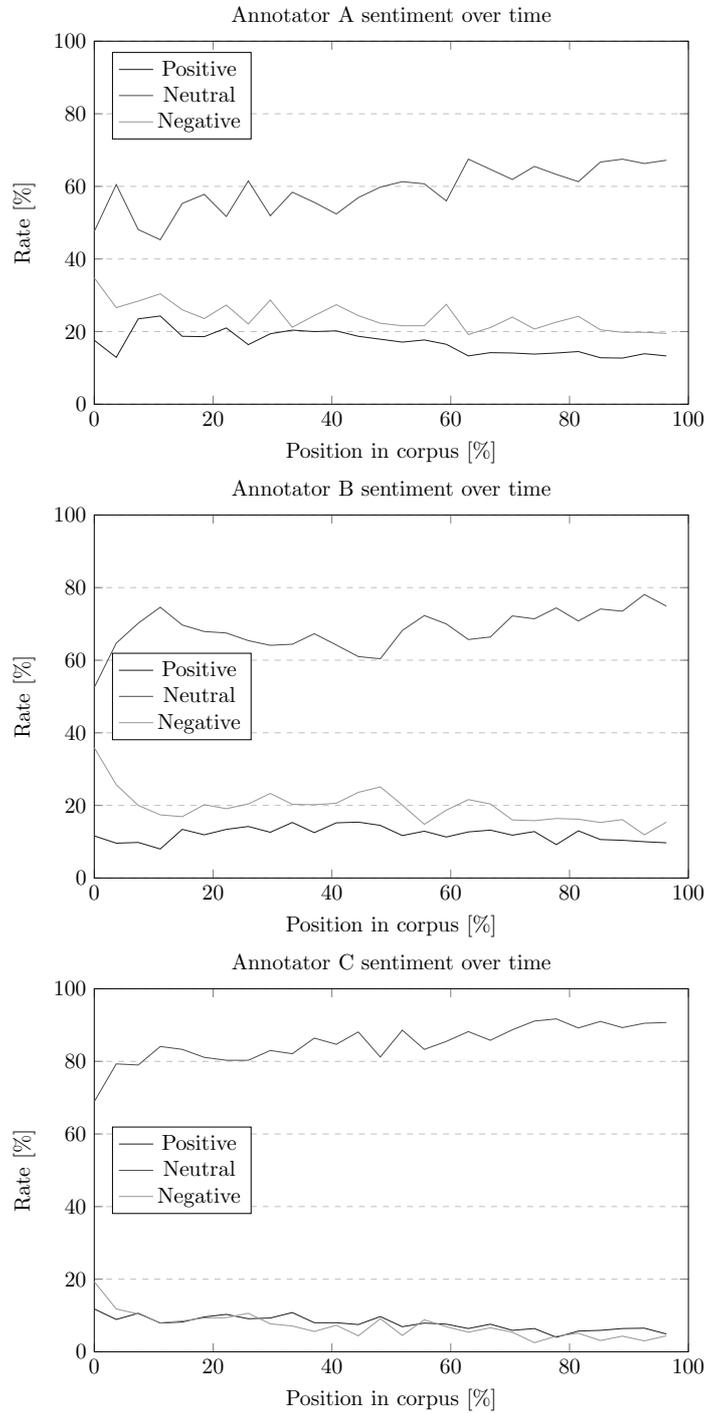
\begin{figure}[htbp]%
\centering
\begin{minipage}{9.5 cm}%
\begin{tikzpicture}[scale=0.75]%
\selectcolormodel{gray}
\begin{axis}[
    title={Annotator A sentiment over time},
    xlabel={Position in corpus [\%]},
    ylabel={Rate [\%]},
    xmin=0, xmax=100,
    ymin=0, ymax=100,
    xtick={0,20,40,60,80,100},
    ytick={0,20,40,60,80,100},
    legend pos=north west,
    ymajorgrids=true,
    grid style=dashed,
]
\addplot[color=blue]
    coordinates {
    (0.0, 17.6)(3.7037037037037037, 12.9)(7.407407407407407, 23.5)(11.11111111111111, 24.3)(14.814814814814815, 18.7)(18.51851851851852, 18.6)(22.22222222222222, 21.0)(25.925925925925927, 16.4)(29.62962962962963, 19.4)(33.333333333333336, 20.4)(37.03703703703704, 20.0)(40.74074074074074, 20.2)(44.44444444444444, 18.7)(48.148148148148145, 17.9)(51.851851851851855, 17.1)(55.55555555555556, 17.7)(59.25925925925926, 16.5)(62.96296296296296, 13.3)(66.66666666666667, 14.2)(70.37037037037037, 14.1)(74.07407407407408, 13.8)(77.77777777777777, 14.1)(81.48148148148148, 14.5)(85.18518518518519, 12.8)(88.88888888888889, 12.7)(92.5925925925926, 13.9)(96.29629629629629, 13.3)
    };
\addlegendentry{Positive}
\addplot[color=red]
    coordinates {
    (0.0, 47.6)(3.7037037037037037, 60.5)(7.407407407407407, 48.1)(11.11111111111111, 45.3)(14.814814814814815, 55.3)(18.51851851851852, 57.8)(22.22222222222222, 51.7)(25.925925925925927, 61.5)(29.62962962962963, 51.9)(33.333333333333336, 58.4)(37.03703703703704, 55.6)(40.74074074074074, 52.4)(44.44444444444444, 56.9)(48.148148148148145, 59.8)(51.851851851851855, 61.3)(55.55555555555556, 60.7)(59.25925925925926, 56.0)(62.96296296296296, 67.5)(66.66666666666667, 64.7)(70.37037037037037, 61.9)(74.07407407407408, 65.5)(77.77777777777777, 63.3)(81.48148148148148, 61.3)(85.18518518518519, 66.7)(88.88888888888889, 67.5)(92.5925925925926, 66.3)(96.29629629629629, 67.2)
    };
\addlegendentry{Neutral}
\addplot[color=green]
    coordinates {
    (0.0, 34.8)(3.7037037037037037, 26.6)(7.407407407407407, 28.4)(11.11111111111111, 30.4)(14.814814814814815, 26.0)(18.51851851851852, 23.6)(22.22222222222222, 27.3)(25.925925925925927, 22.1)(29.62962962962963, 28.7)(33.333333333333336, 21.2)(37.03703703703704, 24.4)(40.74074074074074, 27.4)(44.44444444444444, 24.4)(48.148148148148145, 22.3)(51.851851851851855, 21.6)(55.55555555555556, 21.6)(59.25925925925926, 27.5)(62.96296296296296, 19.2)(66.66666666666667, 21.1)(70.37037037037037, 24.0)(74.07407407407408, 20.7)(77.77777777777777, 22.6)(81.48148148148148, 24.2)(85.18518518518519, 20.5)(88.88888888888889, 19.8)(92.5925925925926, 19.8)(96.29629629629629, 19.5)
    };
\addlegendentry{Negative}
\end{axis}
\end{tikzpicture}
\end{minipage}

\begin{minipage}{9.5 cm}
\begin{tikzpicture}[scale=0.75]
\selectcolormodel{gray}
\begin{axis}[
    title={Annotator B sentiment over time},
    xlabel={Position in corpus [\%]},
    ylabel={Rate [\%]},
    xmin=0, xmax=100,
    ymin=0, ymax=100,
    xtick={0,20,40,60,80,100},
    ytick={0,20,40,60,80,100},
    legend style={at={(0.03,0.5)},anchor=west},
    ymajorgrids=true,
    grid style=dashed,
]
\addplot[color=blue]
    coordinates {
    (0.0, 11.6)(3.7037037037037037, 9.6)(7.407407407407407, 9.8)(11.11111111111111, 8.0)(14.814814814814815, 13.4)(18.51851851851852, 11.9)(22.22222222222222, 13.4)(25.925925925925927, 14.2)(29.62962962962963, 12.6)(33.333333333333336, 15.3)(37.03703703703704, 12.5)(40.74074074074074, 15.2)(44.44444444444444, 15.4)(48.148148148148145, 14.5)(51.851851851851855, 11.7)(55.55555555555556, 12.9)(59.25925925925926, 11.3)(62.96296296296296, 12.7)(66.66666666666667, 13.2)(70.37037037037037, 11.8)(74.07407407407408, 12.8)(77.77777777777777, 9.2)(81.48148148148148, 13.0)(85.18518518518519, 10.6)(88.88888888888889, 10.4)(92.5925925925926, 10.0)(96.29629629629629, 9.7)
    };
\addlegendentry{Positive}
\addplot[color=red]
    coordinates {
    (0.0, 52.5)(3.7037037037037037, 64.7)(7.407407407407407, 70.2)(11.11111111111111, 74.6)(14.814814814814815, 69.7)(18.51851851851852, 67.9)(22.22222222222222, 67.5)(25.925925925925927, 65.4)(29.62962962962963, 64.1)(33.333333333333336, 64.4)(37.03703703703704, 67.3)(40.74074074074074, 64.2)(44.44444444444444, 61.0)(48.148148148148145, 60.4)(51.851851851851855, 68.2)(55.55555555555556, 72.3)(59.25925925925926, 70.0)(62.96296296296296, 65.7)(66.66666666666667, 66.4)(70.37037037037037, 72.2)(74.07407407407408, 71.4)(77.77777777777777, 74.4)(81.48148148148148, 70.8)(85.18518518518519, 74.1)(88.88888888888889, 73.5)(92.5925925925926, 78.1)(96.29629629629629, 74.9)
    };
\addlegendentry{Neutral}
\addplot[color=green]
    coordinates {
    (0.0, 35.9)(3.7037037037037037, 25.7)(7.407407407407407, 20.0)(11.11111111111111, 17.4)(14.814814814814815, 16.9)(18.51851851851852, 20.2)(22.22222222222222, 19.1)(25.925925925925927, 20.4)(29.62962962962963, 23.3)(33.333333333333336, 20.3)(37.03703703703704, 20.2)(40.74074074074074, 20.6)(44.44444444444444, 23.6)(48.148148148148145, 25.1)(51.851851851851855, 20.1)(55.55555555555556, 14.8)(59.25925925925926, 18.7)(62.96296296296296, 21.6)(66.66666666666667, 20.4)(70.37037037037037, 16.0)(74.07407407407408, 15.8)(77.77777777777777, 16.4)(81.48148148148148, 16.2)(85.18518518518519, 15.3)(88.88888888888889, 16.1)(92.5925925925926, 11.9)(96.29629629629629, 15.4)
    };
\addlegendentry{Negative}
\end{axis}
\end{tikzpicture}
\end{minipage}

\begin{minipage}{9.5 cm}
\begin{tikzpicture}[scale=0.75]
\selectcolormodel{gray}
\begin{axis}[
    title={Annotator C sentiment over time},
    xlabel={Position in corpus [\%]},
    ylabel={Rate [\%]},
    xmin=0, xmax=100,
    ymin=0, ymax=100,
    xtick={0,20,40,60,80,100},
    ytick={0,20,40,60,80,100},
    legend style={at={(0.03,0.5)},anchor=west},
    ymajorgrids=true,
    grid style=dashed,
]
\addplot[color=blue]
    coordinates {
    (0.0, 11.8)(3.7037037037037037, 8.9)(7.407407407407407, 10.6)(11.11111111111111, 7.9)(14.814814814814815, 8.2)(18.51851851851852, 9.6)(22.22222222222222, 10.3)(25.925925925925927, 9.1)(29.62962962962963, 9.3)(33.333333333333336, 10.8)(37.03703703703704, 8.0)(40.74074074074074, 8.0)(44.44444444444444, 7.5)(48.148148148148145, 9.7)(51.851851851851855, 6.9)(55.55555555555556, 7.9)(59.25925925925926, 7.6)(62.96296296296296, 6.4)(66.66666666666667, 7.6)(70.37037037037037, 5.9)(74.07407407407408, 6.4)(77.77777777777777, 4.0)(81.48148148148148, 5.7)(85.18518518518519, 5.9)(88.88888888888889, 6.4)(92.5925925925926, 6.5)(96.29629629629629, 4.9)
    };
\addlegendentry{Positive}
\addplot[color=red]
    coordinates {
    (0.0, 68.9)(3.7037037037037037, 79.3)(7.407407407407407, 79.0)(11.11111111111111, 84.1)(14.814814814814815, 83.3)(18.51851851851852, 81.1)(22.22222222222222, 80.3)(25.925925925925927, 80.3)(29.62962962962963, 83.0)(33.333333333333336, 82.1)(37.03703703703704, 86.4)(40.74074074074074, 84.7)(44.44444444444444, 88.1)(48.148148148148145, 81.2)(51.851851851851855, 88.6)(55.55555555555556, 83.3)(59.25925925925926, 85.5)(62.96296296296296, 88.2)(66.66666666666667, 85.8)(70.37037037037037, 88.7)(74.07407407407408, 91.1)(77.77777777777777, 91.7)(81.48148148148148, 89.2)(85.18518518518519, 91.0)(88.88888888888889, 89.3)(92.5925925925926, 90.5)(96.29629629629629, 90.7)
    };
\addlegendentry{Neutral}
\addplot[color=green]
    coordinates {
    (0.0, 19.3)(3.7037037037037037, 11.8)(7.407407407407407, 10.4)(11.11111111111111, 8.0)(14.814814814814815, 8.5)(18.51851851851852, 9.3)(22.22222222222222, 9.4)(25.925925925925927, 10.6)(29.62962962962963, 7.7)(33.333333333333336, 7.1)(37.03703703703704, 5.6)(40.74074074074074, 7.3)(44.44444444444444, 4.4)(48.148148148148145, 9.1)(51.851851851851855, 4.5)(55.55555555555556, 8.8)(59.25925925925926, 6.9)(62.96296296296296, 5.4)(66.66666666666667, 6.6)(70.37037037037037, 5.4)(74.07407407407408, 2.5)(77.77777777777777, 4.3)(81.48148148148148, 5.1)(85.18518518518519, 3.1)(88.88888888888889, 4.3)(92.5925925925926, 3.0)(96.29629629629629, 4.4)
    };
\addlegendentry{Negative}
\end{axis}
\end{tikzpicture}
\end{minipage}

\caption{Annotator-assigned sentiment over time}%
\label{fig:sentiment-over-time}
\end{figure}

In Table~\ref{tab:annotation-distribution}, we see the corpus distribution of perceived sentence polarity for each annotator. Both annotators A and B find more negative than positive statements whereas annotator C finds a roughly equal amount of them. In Figure~\ref{fig:sentiment-over-time}, we see a tendency that is consistent for all three annotators over time, i.e. the number of statements perceived to be neutral grows towards the end of the task, but their ratio of positive vs. negative remains largely the same. 

\begin{table}[h]
\centering
\begin{tabular}{|c|c|c|c|c|}
\hline
     Annotator & Positive & Neutral & Negative & Pos-neg ratio \\
     \hline
     \textbf{A}& 4,576 (17.0\%) & 15,927 (59.0\%) & 6,497 (24.1\%) & 70.4\% \\
     \textbf{B}& 3,267 (12.1\%)& 18,459 (68.4\%) & 5,274 (19.5\%) & 61.9\% \\

     \textbf{C}& 2,118 (7.8\%) & 22,954 (85.0\%) & 1,928 (7.1\%) & 109.9\% \\
     \textbf{Average}& 3,320 (12.3\%) & 19,113 (70.8\%) & 4,566 (16.9\%) & 72.2\%\\
     \hline
\end{tabular}
\caption{Distribution of annotations}
\label{tab:annotation-distribution}
\end{table}

\subsection{Inter-Annotator Agreement}

We computed \emph{agreement}, i.e. how often annotators made the same annotation, \emph{strong disagreement}, i.e. how often one annotator annotated a sentence as positive and another as negative, and \emph{Krippendorff's alpha}  \citep{krippendorff2011computing}.

Krippendorff's alpha is convenient because it generalises to scoring the agreement between more than two annotators. 
Because the human annotators had the task of making a categorical judgement, rather than using a finer scale, we have used the nominal level of measurement in calculating Krippendorff's alpha, meaning that all disagreements have the same weight, whether between negative and neutral or between negative and positive.

In Table~\ref{tab:coincidence}, we see how the annotators agreed, on the data set level.

\begin{table}[h]
\centering
\begin{tabular}{|r|c|c|c|}
\hline
& \multicolumn{3}{c|}{A} \\
& \textbf{POS} & \textbf{NEUTR} & \textbf{NEG} \\
\hline
\textbf{POS} & 2,651 & 552 & 64 \\
B \textbf{NEUTR} & 1,621  & 14,109 & 2,729  \\
\textbf{NEG} & 304 &  1,266 & 3,704 \\
     \hline
\end{tabular}

\begin{tabular}{|r|c|c|c|}
\hline
& \multicolumn{3}{c|}{A} \\
& \textbf{POS} & \textbf{NEUTR} & \textbf{NEG} \\
\hline
\textbf{POS} & 1,868 & 133 & 177 \\
C \textbf{NEUTR} & 2,631  & 15,571  & 4,752  \\
\textbf{NEG} & 77 & 223 & 1,628 \\
     \hline
\end{tabular}

\begin{tabular}{|r|c|c|c|}
\hline
& \multicolumn{3}{c|}{B} \\
& \textbf{POS} & \textbf{NEUTR} & \textbf{NEG} \\
\hline
\textbf{POS} & 1,619 & 310 & 189 \\
C \textbf{NEUTR} & 1,641 & 17,779 & 3,534  \\
\textbf{NEG} & 7 & 370 & 1,551 \\
     \hline
\end{tabular}

\caption{Coincidence matrix of annotator pairs}
\label{tab:coincidence}
\end{table}

In Table~\ref{tab:agreement}, we calculated the agreement, strong agreement and Krippendorf's alpha between the annotators on the data set level.

\begin{table}[h]
\centering
\begin{tabular}{|c|c|c|c|}
\hline
Annotators & Agreement & Strong disagreement & Krippendorff's alpha \\
\hline
\textbf{A} and \textbf{B} & 20,464 (75.8\%)&  368 (1.4\%)&  0.54 \\
\textbf{A} and \textbf{C} & 19,067 (70.6\%) & 194 (0.7\%) & 0.34 \\ 
\textbf{B} and \textbf{C} & 20.949 (77.6\%) & 196 (0.7\%) & 0.44 \\ 
\textbf{A}, \textbf{B} and \textbf{C} & 16,866 (62.5\%) & 505 (1.9\%) & 0.44\\
\hline
\end{tabular}
\caption{Annotator agreements}
\label{tab:agreement}
\end{table}

Out of the 505 instances of strong disagreement among human annotators, 252 were cases where each of the three possible annotations was selected by an annotator, meaning that in these cases there was no majority opinion.

\subsection{Inter-Annotator Agreement Timeline}

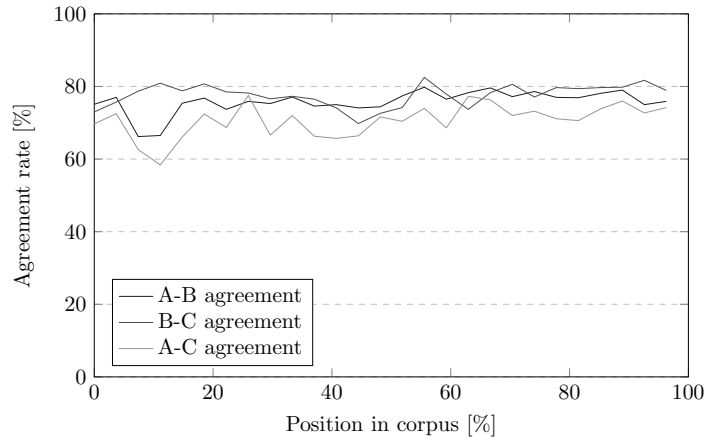
\begin{figure}[htbp]
\centering
\begin{minipage}{9.5 cm}
\begin{tikzpicture}[scale=0.75]
\selectcolormodel{gray}
\begin{axis}[
    xlabel={Position in corpus [\%]},
    ylabel={Agreement rate [\%]},
    xmin=0, xmax=100,
    ymin=0, ymax=100,
    xtick={0,20,40,60,80,100},
    ytick={0,20,40,60,80,100},
    legend pos=south west,
    ymajorgrids=true,
    grid style=dashed,
]

\addplot[color=blue]
    coordinates {
    (0.0, 75.1)(3.7037037037037037, 77.0)(7.407407407407407, 66.2)(11.11111111111111, 66.5)(14.814814814814815, 75.4)(18.51851851851852, 76.8)(22.22222222222222, 73.7)(25.925925925925927, 75.9)(29.62962962962963, 75.3)(33.333333333333336, 77.1)(37.03703703703704, 74.6)(40.74074074074074, 75.0)(44.44444444444444, 74.1)(48.148148148148145, 74.4)(51.851851851851855, 77.4)(55.55555555555556, 79.8)(59.25925925925926, 76.5)(62.96296296296296, 78.3)(66.66666666666667, 79.6)(70.37037037037037, 77.2)(74.07407407407408, 78.6)(77.77777777777777, 77.0)(81.48148148148148, 76.9)(85.18518518518519, 78.1)(88.88888888888889, 79.0)(92.5925925925926, 75.0)(96.29629629629629, 75.9)
    };
\addlegendentry{A-B agreement}

\addplot[color=red]
    coordinates {
    (0.0, 73.0)(3.7037037037037037, 75.7)(7.407407407407407, 78.7)(11.11111111111111, 80.9)(14.814814814814815, 78.8)(18.51851851851852, 80.7)(22.22222222222222, 78.5)(25.925925925925927, 78.2)(29.62962962962963, 76.6)(33.333333333333336, 77.3)(37.03703703703704, 76.5)(40.74074074074074, 74.1)(44.44444444444444, 69.8)(48.148148148148145, 72.6)(51.851851851851855, 74.2)(55.55555555555556, 82.5)(59.25925925925926, 77.9)(62.96296296296296, 73.7)(66.66666666666667, 78.3)(70.37037037037037, 80.6)(74.07407407407408, 77.1)(77.77777777777777, 79.7)(81.48148148148148, 79.4)(85.18518518518519, 79.7)(88.88888888888889, 79.8)(92.5925925925926, 81.7)(96.29629629629629, 78.9)
    };
\addlegendentry{B-C agreement}

\addplot[color=green]
    coordinates {
    (0.0, 69.7)(3.7037037037037037, 72.5)(7.407407407407407, 62.6)(11.11111111111111, 58.4)(14.814814814814815, 66.1)(18.51851851851852, 72.4)(22.22222222222222, 68.7)(25.925925925925927, 77.5)(29.62962962962963, 66.6)(33.333333333333336, 72.0)(37.03703703703704, 66.3)(40.74074074074074, 65.7)(44.44444444444444, 66.4)(48.148148148148145, 71.6)(51.851851851851855, 70.4)(55.55555555555556, 74.0)(59.25925925925926, 68.6)(62.96296296296296, 77.3)(66.66666666666667, 76.3)(70.37037037037037, 72.0)(74.07407407407408, 73.2)(77.77777777777777, 71.1)(81.48148148148148, 70.6)(85.18518518518519, 73.8)(88.88888888888889, 76.0)(92.5925925925926, 72.7)(96.29629629629629, 74.2)
    };
\addlegendentry{A-C agreement}
    
\end{axis}
\end{tikzpicture}
\end{minipage}
\caption{Annotator agreement over time}
\label{fig:agreement-over-time}
\end{figure}

Figure~\ref{fig:agreement-over-time} shows how the inter-annotator agreement developed over time. 
When more than half of the corpus had been annotated, there seems to be more agreement between the annotators whereas their agreement on sentence polarity is less in the initial part of the corpus.

\subsection{Some example annotations}

To illustrate the content of the corpus and the task that the annotators were faced with, we provide some examples from the corpus of some cases we consider indicative of non-obvious choices made by the annotators.

All human annotators tended to agree on a positive sentiment when the sentence contained only a positive assessment of something, whether the commentator's mood, some topic of conversation, or another commentator, even if the sentiment was only a minor part of the comment:

\begin{displayquote}
``no mielestäni kuulostat mielenkiintoiselta, olen itse samankaltaisista asioista kiinnostunut nainen, en pidä baareista, kesällä kun on vapaata olen mieluummin puistossa tai rannalla, mutta puistoista lähden sitten siinä vaiheessa kun muut tulevat sinne ryyppäämään."\vspace{4pt}

\emph{Well, I think you sound interesting, I'm a woman interested in similar things, I don't like bars, in the summer when I have spare time I prefer to spend time in a park or on the beach, but I leave the parks when other people get there to booze.}\vspace{4pt}

\textbf{A pos, B pos, C pos}
\end{displayquote}

Annotators also agreed on the positive sentiment of sentences in cases where there was a clear and unambiguous expression of tone, by using words indicating politeness or smiley faces. Eg:

\begin{displayquote}
``Kiitos kaikille vastaajille!"\vspace{4pt}

\emph{Thanks to everyone who replied!}\vspace{4pt}

\textbf{A pos, B pos, C pos}
\end{displayquote}

Here is a positively annotated case with no explicitly positive content, but which is conciliatory in tone:

\begin{displayquote}
``Itse asiassa pystymetsäläiset ja kruunuhakalaiset on ihan yhtä hyvää jengiä, ei tee tiukkaa."\vspace{4pt}

\emph{Actually people from the countryside and the city are just as good people, no doubt.}\vspace{4pt}

\textbf{A pos, B pos, C pos}
\end{displayquote}

This direct statement of the commentator's own satisfaction with his situation was annotated as positive:

\begin{displayquote}

``Joo kyllä itse olen ihan tyytyväinen palkkaani."\vspace{4pt}

\emph{Yeah, I'm quite satisfied with my salary.}\vspace{4pt}

\textbf{A pos, B pos, C pos}
\end{displayquote}

Negative mood, even when not directly indicating sentiment, was annotated as negative, as in the following example which all human annotators marked as negative:

\begin{displayquote}
``Nuku hyvin, Viivuska :'( \ensuremath\heartsuit"\vspace{4pt}

\emph{Sleep well, Viivuska :'( \ensuremath\heartsuit}\vspace{4pt}

\textbf{A neg, B neg, C neg}
\end{displayquote}

This comment indicating that an argument is taking place was annotated as negative:

\begin{displayquote}
``Missä kohtaa olen sinua nimitellyt?"\vspace{4pt}

\emph{Where exactly did I call you names?}\vspace{4pt}

\textbf{A neg, B neg, C neg}
\end{displayquote}

Some annotations, such as this negative one, require considerable knowledge about the world to interpret and assess:

\begin{displayquote}
``Vihreä puolue ei ole edustanut vihreitä arvoja enää ainakaan puoleen vuosikymmeneen."\vspace{4pt}

\emph{The green party hasn't represented green values for at least half a decade.}\vspace{4pt}

\textbf{A neg, B neg, C neg}
\end{displayquote}

Annotators selected differing annotations especially in cases where multiple sentiments were expressed, as in this case where each of positive, negative and neutral sentiments were selected:

\begin{displayquote}
``Haastattelu meni tosi hyvin ja portfolioon olen panostanut paljon mutta en siltikään usko että pääsen koska en ole käynyt lukiota eikä ne mielellään ota meikäläisiä :/"\vspace{4pt}

\emph{The interview went really well and I put a lot of work into my portfolio but I still don't think I'll get in because I didn't go to secondary school and they don't like to pick people like us :/}\vspace{4pt}

\textbf{A pos, B neg, C neu}
\end{displayquote}

\section{Data Set}
\label{sec:dataset}

Based on the annotations we had obtained, we proceeded to create two gold standard data sets of the annotations. 
One took the majority vote of the annotations and the other derived a 5-grade scale that is often used in shared tasks.

\subsection{Majority vote}

\begin{figure}[ht]
\centering
\begin{minipage}{9.5 cm}
\begin{tikzpicture}[scale=0.75]
\selectcolormodel{gray}
\begin{axis}[
    xlabel={Position in corpus [\%]},
    ylabel={Rate [\%]},
    xmin=0, xmax=100,
    ymin=0, ymax=100,
    xtick={0,20,40,60,80,100},
    ytick={0,20,40,60,80,100},
    legend pos=north west,
    ymajorgrids=true,
    grid style=dashed,
]
\addplot[color=blue]
    coordinates {
    (0.0, 12.2)(3.7037037037037037, 9.4)(7.407407407407407, 12.3)(11.11111111111111, 9.7)(14.814814814814815, 12.6)(18.51851851851852, 11.8)(22.22222222222222, 13.7)(25.925925925925927, 12.4)(29.62962962962963, 13.4)(33.333333333333336, 14.2)(37.03703703703704, 11.7)(40.74074074074074, 14.2)(44.44444444444444, 12.9)(48.148148148148145, 12.7)(51.851851851851855, 11.7)(55.55555555555556, 12.0)(59.25925925925926, 12.0)(62.96296296296296, 10.3)(66.66666666666667, 11.7)(70.37037037037037, 9.9)(74.07407407407408, 10.8)(77.77777777777777, 9.0)(81.48148148148148, 11.0)(85.18518518518519, 9.0)(88.88888888888889, 9.1)(92.5925925925926, 8.9)(96.29629629629629, 8.0)
    };
\addlegendentry{Positive}
\addplot[color=red]
    coordinates {
    (0.0, 58.2)(3.7037037037037037, 70.1)(7.407407407407407, 70.8)(11.11111111111111, 73.2)(14.814814814814815, 72.1)(18.51851851851852, 71.9)(22.22222222222222, 69.5)(25.925925925925927, 72.1)(29.62962962962963, 68.3)(33.333333333333336, 70.7)(37.03703703703704, 73.0)(40.74074074074074, 68.8)(44.44444444444444, 70.5)(48.148148148148145, 70.4)(51.851851851851855, 74.2)(55.55555555555556, 73.6)(59.25925925925926, 72.1)(62.96296296296296, 75.7)(66.66666666666667, 74.2)(70.37037037037037, 77.3)(74.07407407407408, 78.2)(77.77777777777777, 78.9)(81.48148148148148, 76.6)(85.18518518518519, 80.1)(88.88888888888889, 78.8)(92.5925925925926, 82.8)(96.29629629629629, 80.4)
    };
\addlegendentry{Neutral}
\addplot[color=green]
    coordinates {
    (0.0, 29.6)(3.7037037037037037, 20.5)(7.407407407407407, 16.9)(11.11111111111111, 17.1)(14.814814814814815, 15.3)(18.51851851851852, 16.3)(22.22222222222222, 16.8)(25.925925925925927, 15.5)(29.62962962962963, 18.3)(33.333333333333336, 15.1)(37.03703703703704, 15.3)(40.74074074074074, 17.0)(44.44444444444444, 16.6)(48.148148148148145, 16.9)(51.851851851851855, 14.1)(55.55555555555556, 14.4)(59.25925925925926, 15.9)(62.96296296296296, 14.0)(66.66666666666667, 14.1)(70.37037037037037, 12.8)(74.07407407407408, 11.0)(77.77777777777777, 12.1)(81.48148148148148, 12.4)(85.18518518518519, 10.9)(88.88888888888889, 12.1)(92.5925925925926, 8.3)(96.29629629629629, 11.6)
    };
\addlegendentry{Negative}
\end{axis}
\end{tikzpicture}
\end{minipage}
\caption{Majority vote sentiment over time}
\label{fig:majority-vote-over-time}
\end{figure}
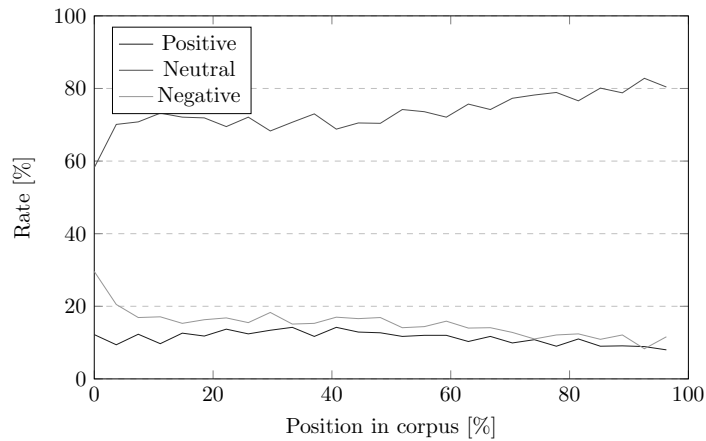

The easiest way to form a gold standard of a polarity annotated corpus of three annotators is to take the majority polarity of the manual annotators, and give a neutral reading for cases where all the annotators disagree. 
The distribution of the majority vote is shown in Table~\ref{tab:majority-vote-distribution} and the distribution over time is shown in Figure~\ref{fig:majority-vote-over-time}.

\begin{table}[h]
\centering
\begin{tabular}{|c|c|}
\hline
      & \#  \\
     \hline
     \textbf{Positive}& 3,066 (11.4\%) \\
     \textbf{Neutral}& 19,825 (73.4\%) \\
     \textbf{Negative}& 4,109 (15.2\%) \\
     \hline
\end{tabular}
\caption{Majority vote distribution}
\label{tab:majority-vote-distribution}
\end{table}

\subsection{Derived Categories (1-5)}

\begin{figure}[htpb]
\centering
\begin{minipage}{9.5 cm}
\begin{tikzpicture}[scale=0.75]
\selectcolormodel{gray}
\begin{axis}[
    xtick={1,2,3,4,5},
    ylabel=Number of sentences,
	xlabel=Derived score,
    width=0.94\textwidth,
    bar width=7pt,
    ybar interval=0.7,
    yticklabel style={
        /pgf/number format/fixed,
        /pgf/number format/precision=5
},
scaled y ticks=false,
	enlargelimits=0.05
]
\addplot+ [ybar interval=0.7
]
	coordinates {(1,1387) (2,6422) (3,14195) (4,3460) (5,1536) (6,1536) };
\end{axis}
\end{tikzpicture}
\end{minipage}
\caption{Derived score sentiment over time}
\label{fig:derived-vote-over-time}
\end{figure}
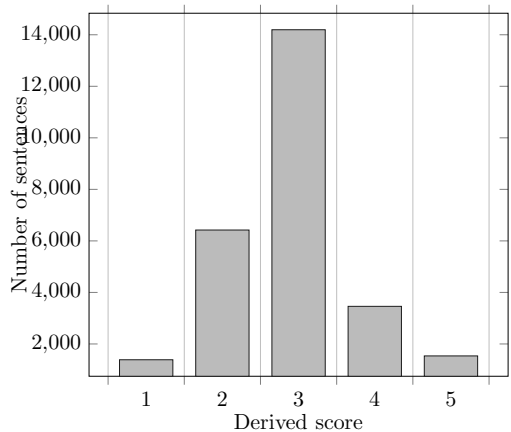

For compatibility with other sources, we also report sentiment on a 1-5 scale for each sentence. 
With $+1$ signifying positive sentiment, $-1$ signifying negative sentiment and $0$ signifying neutral sentiment by a human annotator, we sum the three human scores and map them to the 1-5 scale according to Table~\ref{tab:derived-vote-distribution}. This is illustrated in Figure~\ref{fig:derived-vote-over-time}.

\begin{table}[h]
\centering
\begin{tabular}{c|c|c}
\hline
Sum of evaluations in this corpus & Derived category & Number in corpus \\
\hline
     $-3$ & 1 & 1,387 (5.1\%)\\
     $-2$ or $-1$ & 2 & 6,422 (23.8\%) \\
    $0$ & 3 & 14,195 (52.6\%) \\
    $1$ or $2$ & 4 & 3,460 (12.8\%) \\
    $3$ & 5 & 1,536 (5.7\%)
\end{tabular}
\caption{Derived score distribution}
\label{tab:derived-vote-distribution}
\end{table}

\subsection{Splitting the data set}

We created a split of the data to enable a 20-fold cross-validation corresponding to randomly shuffling the sentences and splitting them into 20 equally-sized portions. 
In each validation run, a different 5\% section can be used for testing, another for development and the remaining 90\% as training data. 
In the gold standard data file, we indicate which split each sentence ended up in for comparability with our test results.
If a cross-validation with fewer splits is preferred, one can simply use several splits for testing and development and the remaining portions for training. 

\subsection{File Format}

The corpus is available in a utf-8 encoded TSV (tab-separated values) file with columns as indicated in Table~\ref{tab:fileformat}. In the table, \emph{split} refers to the cross-validation split to which a sentence belongs, and \emph{batch} to the work package the sentence belongs to. Indexes to the original corpus are strings consisting of a filename, like \verb+comments2008c.vrt+, a space character, and a sentence id number in the file.

\begin{table}[h]
    \centering
\begin{tabular}{|c|c|c|}
\hline
\textbf{Column \#} & \textbf{Column name} & \textbf{Range / data type}\\
\hline
1 & A sentiment & $[-1, 1]$\\
2 & B sentiment & $[-1, 1]$ \\
3 & C sentiment & $[-1, 1]$ \\
4 & majority value & $[-1, 1]$ \\
5 & derived value & $[1, 5]$ \\
6 & pre-annotated sentiment smiley & $[-1, 1]$ \\
7 & pre-annotated sentiment product review & $[-1, 1]$ \\
8 & split \# & $[1, 20]$ \\
9 & batch \# & $[1,9]$ \\
10 & index in original corpus & Filename \& sentence id \\
11 & sentence text & Raw string \\
\hline
\end{tabular}
    \caption{Data set format}
    \label{tab:fileformat}
\end{table}

\section{Initial Experiments with the data set}
\label{sec:experiments}

To evaluate the usefulness of the gold standard data set with a majority vote and the derived scores of the manually annotated corpus, we tested the data set with SentiStrength \citep{thelwall2010sentiment} which is a lexicon-based sentiment analysis program using word lists for various languages. It also has word lists for Finnish. To evaluate the performance of our baseline CNN architecture on different splits of the data set, we used the 20-fold cross-validation split to train separate models.

\subsection{Evaluation Measures}

As evaluation measures, we use agreement, strong agreement and Krippendorf's alpha as indicators of inter-annotator agreement.  

\subsection{Testing a lexicon-based model}

We obtained the SentiStrength \citep{thelwall2010sentiment} rule-based sentiment analysis program and word lists for analysing Finnish texts from its authors. 
It provides for each sentence both a positive and negative sentiment score between $1$ and $5$. 
Taking $score = score_{positive} - score_{negative}$, we convert between scales to be compatible with the majority vote and the derived score. 
The conversion to polarity sentiment is shown in Table~\ref{tab:sentistrength-polarity-conversion}.
 
\begin{table}[h]
\centering
\begin{tabular}{c|c}
     $score$ & Polarity sentiment \\
     \hline
     $< 0$ & Negative \\
     $0$ & Neutral \\
     $> 0$ & Positive
\end{tabular}
\caption{Sentistrength conversion to polarity sentiment}
\label{tab:sentistrength-polarity-conversion}
\end{table}

\begin{figure}[h]
\centering
\begin{minipage}{9.5 cm}
\begin{tikzpicture}[scale=0.75]
\selectcolormodel{gray}
\begin{axis}[
    xlabel={Position in corpus [\%]},
    ylabel={Rate [\%]},
    xmin=0, xmax=100,
    ymin=0, ymax=100,
    xtick={0,20,40,60,80,100},
    ytick={0,20,40,60,80,100},
    legend pos=north west,
    ymajorgrids=true,
    grid style=dashed,
]
\addplot[color=blue]
    coordinates {
    (0.0, 24.5)(3.7037037037037037, 25.4)(7.407407407407407, 26.5)(11.11111111111111, 27.8)(14.814814814814815, 28.3)(18.51851851851852, 26.8)(22.22222222222222, 29.9)(25.925925925925927, 23.5)(29.62962962962963, 25.8)(33.333333333333336, 27.3)(37.03703703703704, 28.4)(40.74074074074074, 27.8)(44.44444444444444, 26.4)(48.148148148148145, 26.5)(51.851851851851855, 24.5)(55.55555555555556, 28.0)(59.25925925925926, 26.0)(62.96296296296296, 26.5)(66.66666666666667, 25.7)(70.37037037037037, 26.5)(74.07407407407408, 27.4)(77.77777777777777, 24.5)(81.48148148148148, 29.0)(85.18518518518519, 26.6)(88.88888888888889, 26.3)(92.5925925925926, 25.8)(96.29629629629629, 24.6)
    };
\addlegendentry{Positive}
\addplot[color=red]
    coordinates {
    (0.0, 67.2)(3.7037037037037037, 65.4)(7.407407407407407, 64.9)(11.11111111111111, 62.7)(14.814814814814815, 65.1)(18.51851851851852, 63.3)(22.22222222222222, 62.7)(25.925925925925927, 68.9)(29.62962962962963, 67.1)(33.333333333333336, 65.7)(37.03703703703704, 64.2)(40.74074074074074, 62.6)(44.44444444444444, 64.9)(48.148148148148145, 63.5)(51.851851851851855, 65.4)(55.55555555555556, 65.0)(59.25925925925926, 65.7)(62.96296296296296, 64.3)(66.66666666666667, 64.9)(70.37037037037037, 63.7)(74.07407407407408, 64.4)(77.77777777777777, 67.0)(81.48148148148148, 63.2)(85.18518518518519, 65.9)(88.88888888888889, 66.5)(92.5925925925926, 67.0)(96.29629629629629, 67.4)
    };
\addlegendentry{Neutral}
\addplot[color=green]
    coordinates {
    (0.0, 8.3)(3.7037037037037037, 9.2)(7.407407407407407, 8.6)(11.11111111111111, 9.5)(14.814814814814815, 6.6)(18.51851851851852, 9.9)(22.22222222222222, 7.4)(25.925925925925927, 7.6)(29.62962962962963, 7.1)(33.333333333333336, 7.0)(37.03703703703704, 7.4)(40.74074074074074, 9.6)(44.44444444444444, 8.7)(48.148148148148145, 10.0)(51.851851851851855, 10.1)(55.55555555555556, 7.0)(59.25925925925926, 8.3)(62.96296296296296, 9.2)(66.66666666666667, 9.4)(70.37037037037037, 9.8)(74.07407407407408, 8.2)(77.77777777777777, 8.5)(81.48148148148148, 7.8)(85.18518518518519, 7.5)(88.88888888888889, 7.2)(92.5925925925926, 7.2)(96.29629629629629, 8.0)
    };
\addlegendentry{Negative}
\end{axis}
\end{tikzpicture}
\end{minipage}
\caption{SentiStrength polarity sentiment over time}
\label{fig:sentistrength-polarity-over-time}
\end{figure}
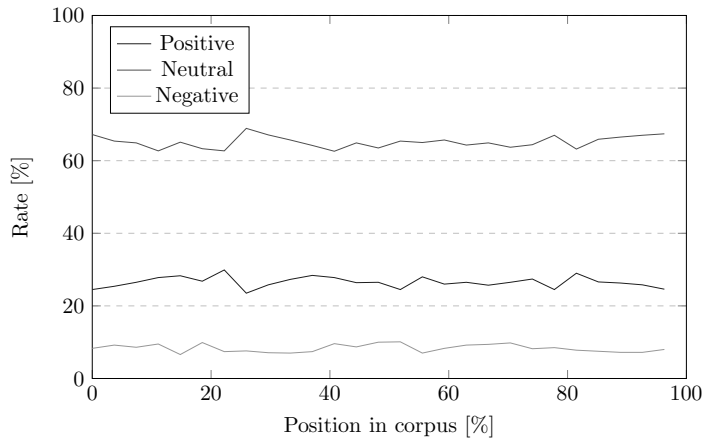

We obtained the results displayed in Table~\ref{tab:sentistrength-polarity-distribution} and illustrated in Figure~\ref{fig:sentistrength-polarity-over-time}.

\begin{table}[h]
\centering
\begin{tabular}{|c|c|c|c|c|}
\hline
     Annotator & Positive & Neutral & Negative & Pos-neg ratio \\
     \hline
     \textbf{SentiStrength}& 7,163 (26.5\%) & 17,586 (65.1\%) & 2,251 (8.3\%) & 318.2\% \\
     \hline
\end{tabular}
\vspace{6pt}

\begin{tabular}{|c|c|c|c|}
\hline
Annotators & Agreement & 
\begin{tabular}{@{}c@{}}Strong \\ disagreement\end{tabular} & \begin{tabular}{@{}c@{}}Krippendorff's \\ alpha\end{tabular} \\
\hline
\textbf{SentiStrength} and Majority vote & 17,248 (63.9\%) & 1,133 (4.2\%) & 0.23 \\
\hline
\end{tabular}
\caption{SentiStrength polarity distribution and majority vote agreement}
\label{tab:sentistrength-polarity-distribution}
\end{table}

The conversion of SentiStrength scores to derived score is shown in Table~\ref{tab:sentistrength-derived-conversion}.

\begin{table}[h]
\centering
\begin{tabular}{c|c}
     $score$ & Derived score \\
     \hline
     $-4 \leq score \leq -3$ & 1 \\
     $-2 \leq score \leq -1$ & 2 \\
     $score = 0$ & 3 \\
     $1 \leq score \leq 2$ & 4 \\
     $3 \leq score \leq 4$ & 5
\end{tabular}
\caption{SentiStrength conversion to derived score}
\label{tab:sentistrength-derived-conversion}
\end{table}

\begin{figure}[h]
\centering
\begin{minipage}{9.5 cm}
\begin{tikzpicture}[scale=0.75]
\selectcolormodel{gray}
\begin{axis}[
    xtick={1,2,3,4,5},
    ylabel=Number of sentences,
	xlabel=Sentistrength derived score,
    width=0.94\textwidth,
    bar width=7pt,
    ybar interval=0.7,
    yticklabel style={
        /pgf/number format/fixed,
        /pgf/number format/precision=5
},
scaled y ticks=false,
	enlargelimits=0.05
]
\addplot+ [ybar interval=0.7
]
	coordinates {(1,368) (2,1883) (3,17586) (4,7015) (5,148) (6,148) };
\end{axis}
\end{tikzpicture}
\end{minipage}
\caption{SentiStrength derived score distribution}
\label{fig:sentistrength-derived-over-time}
\end{figure}
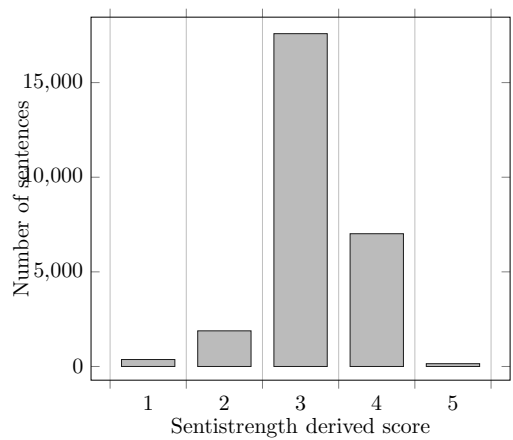

We obtained the results displayed in Table~\ref{tab:sentistrength-derived-distribution} and illustrated in Figure~\ref{fig:sentistrength-derived-over-time}.

\begin{table}[h]
\centering
\begin{tabular}{|c|c|c|c|c|c|}
\hline
Annotator & 1 & 2 & 3 & 4 & 5 \\
\hline
\textbf{Senti-}& 368 & 1,883 & 17,586 & 7,015 & 148 \\
\textbf{Strength} & (1.4\%) & (7.0\%) & (65.1\%) & (26.0\%) & (0.55\%) \\
\hline
\end{tabular}
\vspace{6pt}

\begin{tabular}{|c|c|c|c|}
\hline
Annotators & Agreement & 
\begin{tabular}{@{}c@{}}Agreement with \\ margin of 1\end{tabular} & \begin{tabular}{@{}c@{}}Krippendorff's \\ alpha\end{tabular} \\
\hline
\textbf{SentiStrength} & 13,429 & 2,735 & 0.15 \\
and Derived score & (49.7\%) & (10.1\%) & \\
\hline
\end{tabular}
\caption{SentiStrength derived score distribution and derived score agreement}
\label{tab:sentistrength-derived-distribution}
\end{table}

\subsection{A CNN baseline model}

To evaluate the average performance of the baseline CNN architecture on the data set, we used the 20-fold cross-validation split of the data set to train 20 different CNN models.

In the first model, we used sentences belonging to splits 1 for testing and 2 for development and 3-20 for training. We then gradually shifted the testing and development splits over the whole corpus until we had trained 20 models.

We trained each CNN model with the same architecture as in the preliminary annotations, fitting a mean square error function obtaining the following results, when the regression output value has been scaled to the range $[1, 5]$ and rounded to the nearest integer.

\begin{figure}[h]
\centering
\begin{minipage}{9.5 cm}
\begin{tikzpicture}[scale=0.75]
\selectcolormodel{gray}
\begin{axis}[
    title={CNN 3-class classifier sentiment over time},
    xlabel={Position in corpus [\%]},
    ylabel={Rate [\%]},
    xmin=0, xmax=100,
    ymin=0, ymax=100,
    xtick={0,20,40,60,80,100},
    ytick={0,20,40,60,80,100},
    legend style={at={(0.03,0.5)},anchor=west},
    ymajorgrids=true,
    grid style=dashed,
]
\addplot[color=blue]
    coordinates {
    (0.0, 9.1)(3.7037037037037037, 8.7)(7.407407407407407, 10.0)(11.11111111111111, 8.8)(14.814814814814815, 9.8)(18.51851851851852, 10.8)(22.22222222222222, 9.5)(25.925925925925927, 9.8)(29.62962962962963, 9.1)(33.333333333333336, 10.9)(37.03703703703704, 8.2)(40.74074074074074, 10.3)(44.44444444444444, 9.2)(48.148148148148145, 9.3)(51.851851851851855, 10.4)(55.55555555555556, 9.2)(59.25925925925926, 8.8)(62.96296296296296, 9.4)(66.66666666666667, 10.4)(70.37037037037037, 7.7)(74.07407407407408, 9.1)(77.77777777777777, 9.1)(81.48148148148148, 10.4)(85.18518518518519, 10.4)(88.88888888888889, 9.4)(92.5925925925926, 9.6)(96.29629629629629, 8.5)
    };
\addlegendentry{Positive}
\addplot[color=red]
    coordinates {
    (0.0, 82.1)(3.7037037037037037, 80.9)(7.407407407407407, 79.8)(11.11111111111111, 81.1)(14.814814814814815, 79.6)(18.51851851851852, 79.4)(22.22222222222222, 81.0)(25.925925925925927, 81.2)(29.62962962962963, 81.3)(33.333333333333336, 77.7)(37.03703703703704, 81.0)(40.74074074074074, 80.2)(44.44444444444444, 79.0)(48.148148148148145, 78.2)(51.851851851851855, 78.9)(55.55555555555556, 80.7)(59.25925925925926, 81.2)(62.96296296296296, 80.7)(66.66666666666667, 78.8)(70.37037037037037, 81.8)(74.07407407407408, 80.9)(77.77777777777777, 81.1)(81.48148148148148, 78.8)(85.18518518518519, 80.5)(88.88888888888889, 79.6)(92.5925925925926, 80.6)(96.29629629629629, 80.7)
    };
\addlegendentry{Neutral}
\addplot[color=green]
    coordinates {
    (0.0, 8.8)(3.7037037037037037, 10.4)(7.407407407407407, 10.2)(11.11111111111111, 10.1)(14.814814814814815, 10.6)(18.51851851851852, 9.8)(22.22222222222222, 9.5)(25.925925925925927, 9.0)(29.62962962962963, 9.6)(33.333333333333336, 11.4)(37.03703703703704, 10.8)(40.74074074074074, 9.5)(44.44444444444444, 11.8)(48.148148148148145, 12.5)(51.851851851851855, 10.7)(55.55555555555556, 10.1)(59.25925925925926, 10.0)(62.96296296296296, 9.9)(66.66666666666667, 10.8)(70.37037037037037, 10.5)(74.07407407407408, 10.0)(77.77777777777777, 9.8)(81.48148148148148, 10.8)(85.18518518518519, 9.1)(88.88888888888889, 11.0)(92.5925925925926, 9.8)(96.29629629629629, 10.8)
    };
\addlegendentry{Negative}
\end{axis}
\end{tikzpicture}
\end{minipage}
\caption{CNN polarity sentiment over time}
\label{fig:cnn-polarity-over-time}
\end{figure}
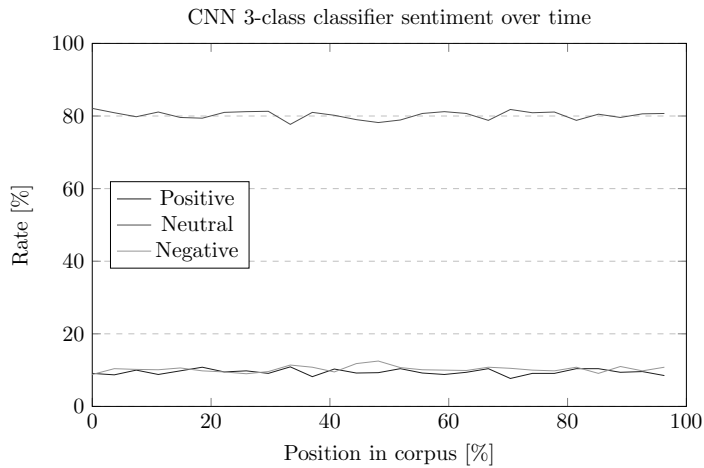

Using the human majority vote in the gold standard data set as training and test data, we obtained the following results for the 20-fold cross-validation as shown in Table~\ref{tab:cnn-polarity-distribution} and illustrated in Figure~\ref{fig:cnn-polarity-over-time}.

\begin{table}[h]
\centering
\begin{tabular}{|c|c|c|c|c|}
\hline
Annotator & Positive & Neutral & Negative & Pos-neg ratio \\
\hline
\begin{tabular}{@{}c@{}}\textbf{CNN 3-class} \\ \textbf{classifier} \end{tabular} & 2,559 (9.5\%) & 21,668 (80.3\%) & 2773 (10.3\%) & 92\% \\
\hline
\end{tabular}
\vspace{6pt}

\begin{tabular}{|c|c|c|c|}
\hline
Annotators & Agreement & 
\begin{tabular}{@{}c@{}}Strong \\ disagreement\end{tabular} & \begin{tabular}{@{}c@{}}Krippendorff's \\ alpha\end{tabular} \\
\hline
\textbf{CNN 3-class classifier} & 16,691 & 658 & 0.45 \\
and Majority vote & (61,8\%) & (2,4\%) &  \\
\hline
\end{tabular}
\caption{CNN polarity distribution and majority vote agreement}
\label{tab:cnn-polarity-distribution}
\end{table}

\begin{figure}[h]
\centering
\begin{minipage}{9.5 cm}
\begin{tikzpicture}[scale=0.75]
\selectcolormodel{gray}
\begin{axis}[
    xtick={1,2,3,4,5},
    ylabel=Number of sentences,
	xlabel=CNN model derived score,
    width=0.94\textwidth,
    bar width=7pt,
    ybar interval=0.7,
    yticklabel style={
        /pgf/number format/fixed,
        /pgf/number format/precision=5
},
scaled y ticks=false,
	enlargelimits=0.05
]
\addplot+ [ybar interval=0.7
]
	coordinates {(1,483) (2,6425) (3,15493) (4,3744) (5,855) (6,855) };
\end{axis}
\end{tikzpicture}
\end{minipage}
\caption{CNN derived score distribution}
\label{fig:cnn-derived-over-time}
\end{figure}
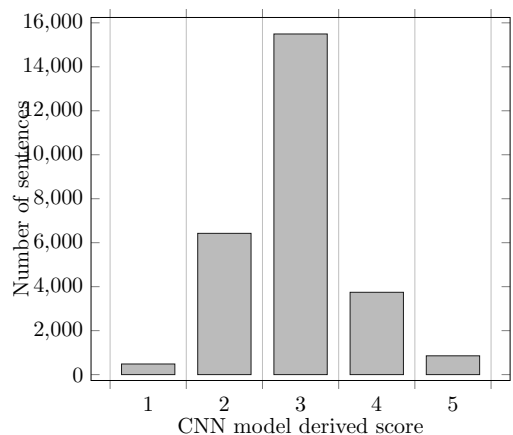

Using the derived score of the gold standard data set as training and test data, we obtained the following results for the 20-fold cross-validation as shown in Table~\ref{tab:cnn-derived-distribution} and illustrated in Figure~\ref{fig:cnn-derived-over-time}. The mean absolute error averaged over all cross-validation runs was 0.54 and the standard deviation was 0.04.

\begin{table}[h]
\centering
\begin{tabular}{|c|c|c|c|c|c|}
\hline
Annotator & 1 & 2 & 3 & 4 & 5 \\
\hline
\textbf{CNN}& 483 & 6,425 & 15,493 & 3,744 & 855 \\
\textbf{architecture}& (1.8\%) & (23.8\%) & (57.4\%) & (13.9\%) & (3.2\%) \\
\hline
\end{tabular}
\vspace{6pt}

\begin{tabular}{|c|c|c|c|}
\hline
Annotators & Agreement & 
\begin{tabular}{@{}c@{}}Agreement with \\ margin of 1\end{tabular} & 
\begin{tabular}{@{}c@{}}Krippendorff's \\ alpha\end{tabular}
\\ 
\hline
\textbf{CNN architecture} & 14,294 & 25,351 & 0.24 \\
and Derived score & (52.9\%) & (90.2\%) &  \\
\hline
\end{tabular}
\caption{CNN derived distribution and derived score agreement}
\label{tab:cnn-derived-distribution}
\end{table}

\subsection{Error Analysis}

A vocabulary-based annotator such as SentiStrength is easily fooled by its inability to detect negation:

\begin{displayquote}
``Mutta ei siellä mitään kamalaa ole!"
\vspace{4pt}

\emph{But there is nothing horrible!}
\vspace{4pt}

\textbf{A pos, B pos, C pos, SentiStrength maximally negative}
\end{displayquote}

Or lacking understanding of compounds, as in this case where it responds to the "horror" in "horror movies":

\begin{displayquote}
``Kiistämättä kyllä parhaita kauhuelokuva aikakausia!"
\vspace{4pt}

\emph{Undeniably one of the best periods for horror movies!}

\textbf{A pos, B pos, C pos, SentiStrength maximally negative}
\end{displayquote}

Errors made by neural networks, such as our baseline CNN model, are harder to interpret, except that they do not appear to be due to the limitation of a finite convolution kernel (up to a maximum of 5 words). 

In the following example, one could even argue that the CNN model is correct and that all three human annotators are wrong, because the underlying sentiment is a sad perhaps even depressed longing to be happy. This example appears to contain the word ``iloinen" \emph{happy} that often appears in positive sentences but has a frowny face:

\begin{displayquote}
``Haluan olla se iloinen tyttö pitkästä aikaa. :("
\vspace{4pt}

\emph{I want to be that happy girl I haven't been for a long time :(}
\vspace{4pt}

\textbf{A pos, B pos, C pos, CNN negative}
\end{displayquote}

The corpus contains quite a few examples of sarcasm and jest, and one sometimes wonders if the CNN models did not in fact get this more often than the human annotators:

\begin{displayquote}
``Äänekäs sovinistiörkki! ;)"
\vspace{4pt}

\emph{You loud chauvinist orc! ;)}
\vspace{4pt}

\textbf{A neg, B neg, C neg, CNN positive}
\end{displayquote}

\section{Discussion and Conclusion}
\label{sec:discussionconclusion}

In our survey of previous work, we noted that there were only two data sets for sentiment analysis of movie subtitles available for Finnish, but no large-scale social media data set with sentiment polarity annotations. 
This publication remedies this short coming by introducing a 27,000-sentence data set annotated independently with sentiment polarity by three native annotators. 
The same three annotators annotated the whole data set. This is in contrast to other data sets which have usually been annotated piecemeal by a large number of annotators. 
Our data set provides a unique opportunity for further studies of annotator behaviour over time, e.g. human inter-annotator agreement seems to increase without coordination.
One can speculate that the annotators become more proficient in their opinion mining towards the end leading to a convergence in their judgements. 
In addition, we test the data set by providing two baselines validating the usefulness of the data set. 
The data set is distributed through the Language Bank of Finland.\footnote{http://urn.fi/urn:nbn:fi:lb-2015120101}

\section{Acknowledgements}
We would like to thank the annotators for their time and FIN-CLARIN and the Language Bank of Finland for access to the data.

\bibliography{Sentiment.bib}  
\bibliographystyle{acl_natbib}

\end{document}